\documentclass[letterpaper, 10 pt, conference]{ieeeconf}  

\IEEEoverridecommandlockouts                              

\usepackage{graphics} 
\usepackage{graphicx,subfigure}
\usepackage{epsfig}
\usepackage[ruled,linesnumbered,vlined]{algorithm2e}
\usepackage{multirow}
\usepackage{caption}
\usepackage{color,soul}
\usepackage{cite}
\title{\LARGE \bf
Gaze Training by Modulated Dropout Improves Imitation Learning
}
\author{Yuying Chen$^{*}$,  \  Congcong Liu$^{*}$,  \  Lei Tai,  \  Ming Liu,  \  Bertram E. Shi  
\thanks{* These two authors contributed equally}
\thanks{This work was supported by the National Natural Science Foundation of China (Grant No. U1713211), the Research Grant Council of Hong Kong SAR Government, China, under Project No. 11210017, No. 21202816 and No. 16213617, awarded to Prof. Ming Liu and Prof. Bert Shi.}
\thanks{All the authors are with the Hong Kong University of Science and Technology. {\tt\small \{ychenco, cliubh, ltai, eelium, eebert\}@ust.hk}}  %
}

\begin{document}

\maketitle
\thispagestyle{empty}
\pagestyle{empty}

\begin{abstract}

Imitation learning by behavioral cloning is a prevalent method that has achieved some success in vision-based autonomous driving.
The basic idea behind behavioral cloning is to have the neural network learn from observing a human expert's behavior.
Typically, a convolutional neural network learns to predict the steering commands from raw driver-view images by mimicking the behaviors of human drivers.
However, there are other cues, such as gaze behavior, available from human drivers that have yet to be exploited.
Previous researches have shown that novice human learners can benefit from observing experts' gaze patterns.
We present here that deep neural networks can also profit from this.
We propose a method, gaze-modulated dropout, for integrating this gaze information into a deep driving network implicitly rather than as an additional input.
Our experimental results demonstrate that gaze-modulated dropout enhances the generalization capability of the network to unseen scenes. Prediction error in steering commands is reduced by 23.5\% compared to uniform dropout. Running closed loop in the simulator, the gaze-modulated dropout net increased the average distance travelled between infractions by 58.5\%.
Consistent with these results, the gaze-modulated dropout net shows lower model uncertainty.
\end{abstract}
\section{Introduction}




End-to-end deep learning has captured much attention and has been widely applied in many autonomous control systems.
Successful attempts have been made in end-to-end driving through imitation learning. 
Among various imitation learning methods, behavioral cloning through supervised learning has been successfully implemented for many tasks including off-road driving and lane following, with both simplicity and efficiency.

Behavioral cloning follows a teacher-student paradigm, where students learn to mimic the teacher's demonstration. Previous behavioral cloning work for autonomous driving mainly focused on learning only the explicit mapping from the sensory input to the control output, with little consideration to other implicit supervisions from teachers despite the fact that there is still a wealth of cues from human experts.

Researches have shown that novice human learners can benefit from observing experts' gaze.
It also holds for novice drivers learning to drive while viewing the expert gaze videos \cite{yamani2017following}. 
Therefore, it is very promising to investigate whether deep driving networks trained by behavioral cloning could benefit from exposure to expert's gaze patterns.

However, it is not yet clear how gaze information can best be integrated into deep neural networks.
Some recent work \cite{liu2019gaze} has incorporated human attention to improve policy performance for learning-based visuomotor tasks. However, they simply add the gaze map as an additional image-like input. We question whether this is the best way to incorporate gaze information. Saccadic eye movements shift the eye gaze in different directions, allocating high resolution processing and attention towards different parts of the visual scene. This suggests that gaze behavior may be better viewed as a modulating effect, rather than as an additional source of information. In addition, treating the gaze map as an additional image input increases the complexity of the network. This additional complexity, however, is inefficiently utilized, as most of the gaze map is close to zero.

To better exploit the information from human eye gaze, which encodes rich information about human attention and intent,
in this paper, we propose gaze-modulated dropout to incorporate gaze information into deep driving networks.
A conditional adversarial network (\textit{Pix2Pix}) is trained to estimate the human eye gaze distribution in the visual scene while driving \cite{liu2019gaze}.
Then, we use the estimated gaze distribution to modulate the dropout probability of units at different spatial locations. Units near the estimated gaze location have a lower dropout probability than units far from the estimated gaze location. We hypothesize that this will help the network focus on task critical objects and ignore task-irrelevant areas such as the background. This should be especially helpful when the network encounters new and unfamiliar environments. In addition, the proposed gaze-modulated dropout does not increase the complexity nor structure of the behavior cloning network. It can be easily inserted into many existing neural network architectures simply by replacing the normal dropout layers.

To validate such a robust generalization ability augmented by gaze-modulated dropout, we propose to evaluate the \textit{epistemic} (model) uncertainty of the trained model.
The \textit{epistemic} uncertainty measures the similarity between newly observed samples and previous observation \cite{mcallister2017concrete}.
 In the context of autonomous driving, the use of \textit{epistemic} uncertainty can help to reveal how well the generalization capability of the driving network is to unseen environments.

The contributions of our work are mainly as follows:
\begin{itemize}
\item We propose gaze-modulated dropout based on the generated gaze maps to incorporate the gaze information into a deep imitation network (\textit{PilotNet}). 
\item We demonstrate that modeling of auxiliary cues not directly related to the commands improves the performance of imitation learning. This takes imitation learning to the next step, by showing the benefits of a more complete understanding of human expert behaviors.
\item We evaluate the model uncertainty of trained models and validate the effectiveness of the imitation network with gaze-modulated dropout, showing significant performance improvements.
\end{itemize}


\section{Related work}\label{relatedwork}


\subsection{End-to-end autonomous driving network}
For vision-based autonomous driving systems, end-to-end methods have become more and more popular as they avoid the decomposition of processes into multiple stages and optimize the whole system simultaneously.

Reinforcement learning algorithms learn to drive in a trial-and-error fashion. There are some works applying deep reinforcement learning to autonomous driving in simulators, such as \textit{Carla} \cite{liang2018cirl, zhang2019vr} and \textit{TORCS} \cite{pan2017virtual}. 
The major limitations are the request of designing a reward function, which can be hard in complex driving tasks, and its low sample efficiency. 

Imitation learning trains an agent from human demonstrations. Bojarski \textit{et al.} \cite{bojarski2016end} trained a convolutional neural network (\textit{PilotNet}) to map visual input to steering commands for road following tasks. 
A similar framework was also applied in an obstacle avoidance task in ground vehicles \cite{muller2006off}. 
Codevilla \textit{et al.} \cite{codevilla2018end} proposed a deep multi-branch imitation network. Directed by high-level commands input from a planning module or human user, it trained a conditional controller to map visual input and speed measurements to action output. The trained controller allowed an autonomous vehicle to stay on the road and to follow the high-level command that represents the expert's intention (go left, go right or go straight). The main limitation of these systems is that they do not generalize well to unseen scenes \cite{liang2018cirl}. For example, the model of \cite{codevilla2018end} trained in Town$1$ of the Carla simulator had obvious performance degradation in Town$2$ even with a number of data augmentation techniques.

As far as we know, no existing systems consider utilizing human gaze behavior in spite of the rich driver-intention-related cues it contains. Furthermore, the uncertainty of the autonomous driving network is rarely considered.

\subsection{Eye gaze in driving}
In the context of a driving scenario, a wealth of cues regarding human intent and decision making are contained in the human eye gaze.
For assisted driving systems, there are works utilizing eye gaze information to monitor the mental state of drivers. 
However, in the field of autonomous driving research, the utilization of eye gaze has not yet been well exploited. Whether and how human gaze can help autonomous driving is still under-explored.

A very related work \cite{palazzi2017predicting} proposed a multi-branch deep neural network to predict eye gaze in urban driving scenarios. It attached much importance to gaze data analysis over different driving scenes and driving conditions. They studied how eye gaze is distributed over different semantic categories in a scene, and the effects of driving speed on driver's attention, to name a few. However, they did not apply the predicted gaze information to an autonomous driving system.


\subsection{Uncertainty in Deep Learning}
The estimation of uncertainty can reflect a model's confidence and what the model does not know \cite{kendall2019geometry}. As a measure of ``familiarity'', \textit{epistemic} uncertainty quantifies how similar a new sample is with previously seen samples \cite{mcallister2017concrete}. Kendall \textit{et al.} \cite{kendall2019geometry} proposed a method to capture \textit{epistemic} uncertainty by sampling over the distribution of model weights using dropout in testing time. This method to quantify the model uncertainty has been applied into various tasks, such as camera relocalization \cite{kendall2016modelling}. They showed that larger \textit{epistemic} uncertainty is observed for objects that are rare in training datasets. The main limitation for this dropout method is that it needs expensive sampling, which is not suitable for real-time applications. However, it can be used as an offline evaluation metric for model confidence to unseen samples.

\section{Methods}\label{method}






\subsection{Framework}
\begin{figure*}[!htb]
  \centering
    {\includegraphics[width=1.75\columnwidth]{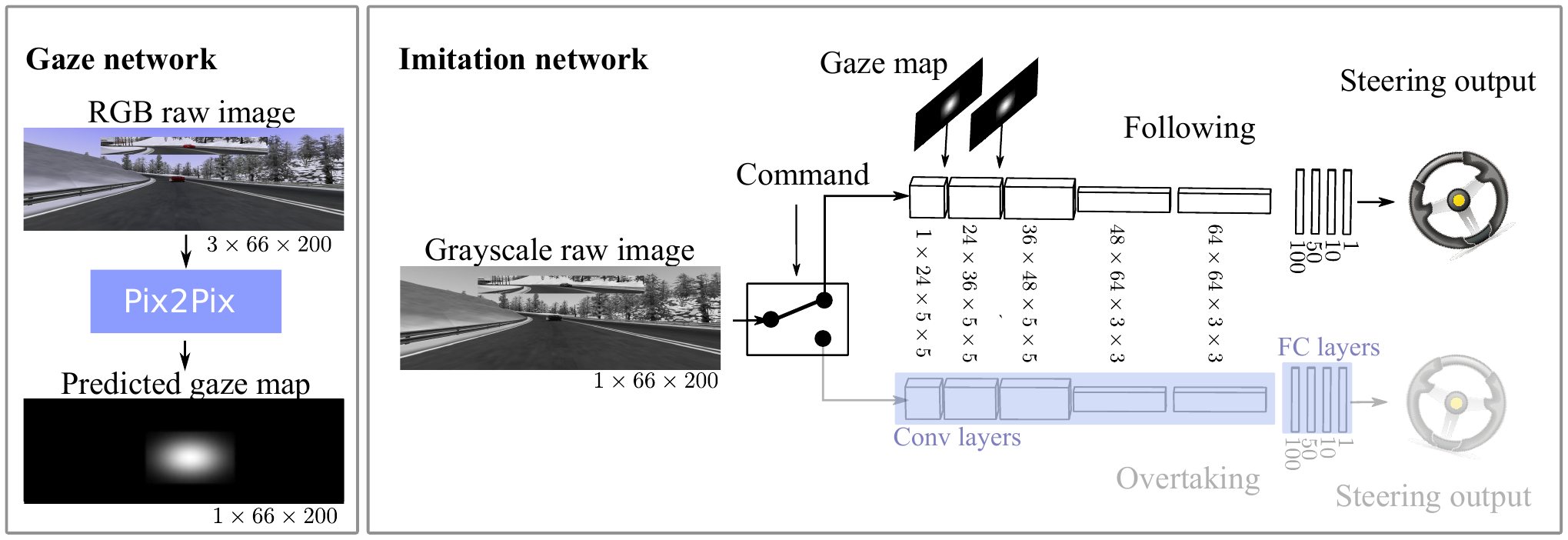}
     \caption{The autonomous driving system with gaze-modulated dropout. The gaze network (\textit{Pix2Pix}) generates a gaze map for the gaze-modulated dropout in the imitation network as per \cite{liu2019gaze}. As the input, the gray-scale drive-view image is forwarded to networks for path following or overtaking according to the command. The filters of convolutional layers appear as blocks marked by corresponding sizes and are flattened to feed into the fully connected (FC) layers. The four vertical bars denote the four FC layers. The final layer has a single scalar unit that encodes the steering command. \label{fig:system}}}
      \vspace{-1.5em}
\end{figure*}

As shown in Fig. \ref{fig:system}, the autonomous driving system with modulated gaze dropout can be divided into two parts: the gaze network and the imitation network.

To achieve the autonomous driving system with gaze information in real time other than a human guided driving system, we synthesized the gaze map by the \textit{Pix2Pix} network as the gaze network following \cite{liu2019gaze}. Trained with pairs of driving scenes and ground truth gaze maps, the gaze network learned to generate gaze maps from the driver-view images.

The generated gaze map was further used as the mask of the dropout, which is referred to as the gaze-modulated dropout. The imitation network has the same structure as \textit{PilotNet} proposed by \cite{bojarski2016end}, which has five convolutional layers and four fully-connected layers. We used ReLU as the activation function, and dropout is applied after the first two convolutional layers.

\subsection{Gaze-modulated dropout}
To help the network focus on task critical objects, we introduced gaze information into the network by dropping fewer units of the highly concerned area of the input images or features and dropping more on the area with low attention.
We refer to this as gaze-modulated dropout.
It is similar to conventional dropout, but with non-uniform drop probability.

\begin{figure}[!htb]
  \centering
    {\includegraphics[width=1.0\columnwidth]{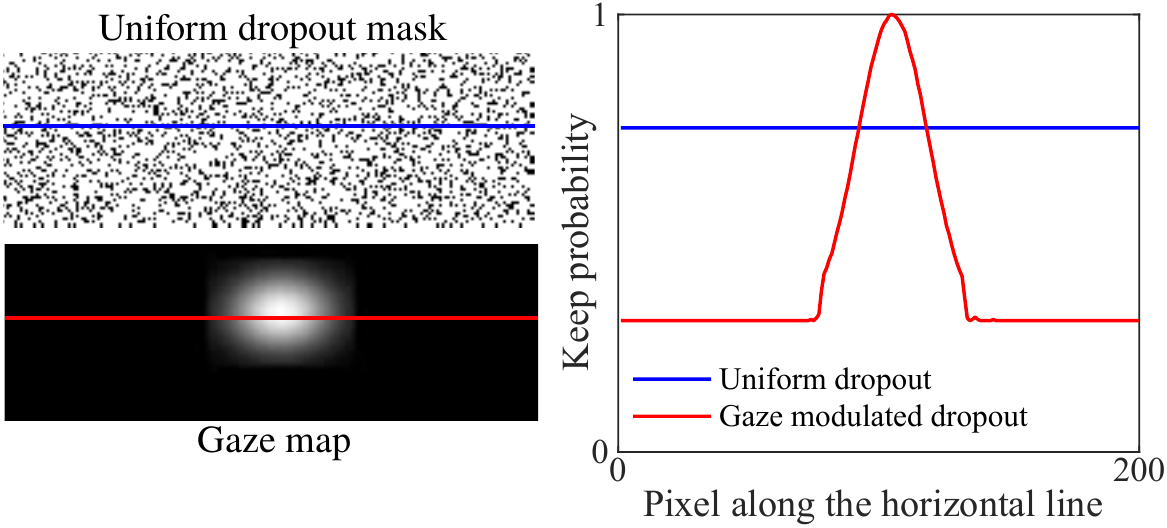}
     \caption{Keep probability settings for gaze-modulated dropout and uniform dropout. \label{fig:dropout}}}
     \vspace{-1em}
\end{figure}

As shown in Fig. \ref{fig:dropout}, the drop probability of gaze-modulated dropout  ($DP_g$) is decided by the gaze map ($\mathbf{G}$).
Along the horizontal line on the gaze map, the keep probability ($KP$) increases when the pixel value increases.
While for the conventional dropout, referred to uniform dropout, the dropout probability ($DP_u$) across the image remains the same. The drop probability of a certain unit with indexes $(i,j)$ can be summarized in (\ref{eqn:uniform}) and (\ref{eqn:gaze}), where $dp$ and $dp_{max}$ are adjustable parameters. $dp$ is the drop probability of all units for uniform dropout, and $dp_{max}$ of gaze-modulated dropout is the maximum drop probability of units corresponding to the zero-value pixels in the gaze map.  As most of the area of the gaze map is zero-value, with slight abuse of notation, we mark $dp_{max}$ as $dp$ for gaze-modulated dropout.
\begin{equation} \label{eqn:uniform}
  DP_{u}(i,j) = dp
\end{equation}
\vspace{-1em}
\begin{equation} \label{eqn:gaze}
  DP_{g}(i,j) = f(\mathbf{G}(i,j),dp_{max} ).
\end{equation}
\vspace{-1em}

The implementation details of gaze map modulated dropout are shown in Algorithm \ref{alg:gaze_drop}. Given the input image or features $\mathbf{F}$ and the gaze map $\mathbf{G}$, a random array $\mathbf{A}$ was generated first with the same width and height as $\mathbf{F}$.
The keep probability mask $\mathbf{K}$ with the same size as $\mathbf{F}$ was first obtained by interpolating the gaze map, and then it was scaled to the range $[1-dp,1]$, with its value representing the keep probability.
By comparing it with $\mathbf{A}$, a binary mask was further generated.
The zero-values of the binary mask set the corresponding pixels of $\mathbf{F}$ to be zero.

For uniform dropout, it is not recommended to apply dropout in testing \cite{srivastava2014dropout} to get an averaged prediction as it requires many thinned models running exponentially.
A simple approximate averaging method was conducted instead, with no dropout at test time.
For gaze-modulated dropout, we applied it in a similar way. It is not feasible to directly remove the gaze-modulated dropout module at the test stage as the gaze information is essential.
For a fair comparison, we kept the gaze-modulated dropout in testing, but also made some modifications to approximate the averaging effect. We studied the effectiveness of the approximation and compared it with the uniform dropout in section \ref{sec:drop_imple}.

\begin{algorithm}
  \SetAlgoLined
  \KwIn{Activation output of a layer ($\mathbf{F}$), gazemap ($\mathbf{G}$), mode, most probability of dropout $dp$}
    $N, H, W, C$ = $shapeof(\mathbf{F}$)\\
    \If{mode==pixel-wise dropout}{  
    Randomly sample array $\mathbf{A}$ with size of $(H, W)$: $\mathbf{A_{i,j}}\sim Uniform(0,1)$ \\
    $KP$ mask $\mathbf{K}$ = $interpolate$($\mathbf{G}$, size=$(H, W)$)\\
    Rescale the range of values of $\mathbf{K}$ to ($1-dp$, 1)\\
    Binary mask $\mathbf{M}$ = $(\mathbf{K}>\mathbf{A})$ \\
    Apply the mask: $\mathbf{F}$ = $\mathbf{F}\times\mathbf{M}$\\

    }
    return $\mathbf{F}$
  \caption{Gaze-modulated dropout\label{alg:gaze_drop}}
\end{algorithm}
\setlength{\textfloatsep}{16pt}

\subsection{Uncertainty}
We evaluated the \textit{epistemic} uncertainty of the trained models, and investigated the effect of gaze-modulated dropout based on the hypothesis that gaze-modulated dropout drops task-irrelevant units and reduces the difference among scenes to improve the generalization capability of the model.

Modelling \textit{epistemic} uncertainty using a stochastic regularization technique like dropout has been proved to be effective for different tasks \cite{gal2016dropout}. By performing forward passes multiple times for the same input, the mean and variance of the output can be obtained. The \textit{epistemic uncertainty} is also captured. In equation (\ref{eqn:mean}) and (\ref{eqn:var}), we show the calculation of the mean and variance of the outputs through multiple forward passes, where $T$ represents the forward times, and $W_t$ represents the masked model weights. For each time $t$, the weights are zeroed with the probability of $dp$ by dropout.
\vspace{-1em}
\begin{equation} \label{eqn:mean}
  \overline{y} = \frac{1}{T}\sum_{t=1}^{T}f^{W_t}(x)
  \end{equation}
  \vspace{-1em}
\begin{equation} \label{eqn:var}
  Var(y) = \frac{1}{T}\sum_{t=1}^{T}(f^{W_t}(x)-\overline{y})^2.
  \end{equation}

\section{Experiments and Results}\label{result}






\subsection{Network training}
\subsubsection{Data collection}

The experiments were conducted in \textit{TORCS} \cite{wymann2000torcs}, which is an open source highway driving simulator. 
We chose \textit{TORCS} as it simulates multi-lane highway diving scenarios, which meet the needs of our task.
For each test, the subject was asked to watch the screen that shows the real-time drive-view scene and control the car with a steering wheel and pedal. For most of the time, lane following was required. Overtaking was executed if it was needed. While the subject was changing lanes, a button was pressed to mark this overtaking maneuver. During the experiments, the gaze data and action data of the subjects were collected. The gaze data was collected by a remote eye tracker, Tobii Pro X60. We chose five different scenes and collected data for four trials in each scene.

\subsubsection{Training and testing details}
For the gaze network, we followed the \textit{Pix2Pix} architecture\cite{isola2017image}. Around 3500 images from $Track1$ and $Track2$ were used for training.
For the imitation network, three trials of $Track1$ and $Track2$, six trials in total, were used for training, which is about 40000 images in total. The remaining one trial in $Track1-2$ and all the trials in $Track3-5$ are used for testing.

For the consideration of robustness, we refer to \cite{codevilla2018end} and add expert demonstrations of recovery from drift. This accounts for a proportion of 10\% of the training data. Furthermore, we performed data augmentations such as random changes in contrast, brightness and gamma online for each image.

We trained two convolutional networks with the same structure but with the following behavior data and overtaking behavior data. During the test, we manually selected the imitation network for path following or overtaking.
\subsubsection{Networks to compare}
We trained the imitation network with uniform dropout as the baseline. This is referred to as \textbf{Uniform dropout} in the results. To evaluate the effectiveness of the gaze-modulated dropout, we implemented three methods.

\textbf{Real gaze-modulated dropout}: The imitation network was trained with dropout given the real gaze map as the mask. To simplify this, the gaze-modulated dropout mentioned below without specification refers to the real gaze-modulated dropout.

\textbf{Estimated gaze-modulated dropout}: The imitation network was trained with dropout given the estimated gaze map as the mask. The estimated gaze map was generated by the gaze network. Under these circumstances, the trained network can be applied to online autonomous driving test.

\textbf{Center Gaussian blob modulated dropout}: The imitation network was trained with dropout given the image with a single Gaussian blob in the center as the mask. The implementation of the Gaussian blob in the image center is based on the observation that the subject mostly looks at the center area of the scene.


\subsection{Gaze Network Evaluation}

\begin{table}[]
\center
\begin{tabular}{ccccccc}
\hline
                    &                                                                 & T1   & T2   & T3   & T4   & T5   \\ \hline
\multirow{2}{*}[-0.7em]{KL} & \begin{tabular}[c]{@{}c@{}}Estimated\\ gaze map\end{tabular}     & 0.75 & 0.63 & 0.94 & 0.88 & 0.82 \\ \cline{2-7}
                    & \begin{tabular}[c]{@{}c@{}}Center \\ Gaussian blob\end{tabular} & 3.08 & 2.57 & 2.63 & 1.95 & 2.37 \\ \hline
\multirow{2}{*}[-0.7em]{CC} & \begin{tabular}[c]{@{}c@{}}Estimated\\ gaze map\end{tabular}     & 0.86 & 0.87 & 0.83 & 0.81 & 0.85 \\ \cline{2-7}
                    & \begin{tabular}[c]{@{}c@{}}Center \\ Gaussian blob\end{tabular} & 0.68 & 0.72 & 0.73 & 0.80 & 0.75 \\ \hline
\end{tabular}
\caption{The similarity of gaze map estimates to the real gaze map over different tracks. KL denotes Kullback-Leibler divergence, and CC denotes Correlation Coefficient.\label{tab:gaze_pred}}
\vspace{-1em}
\end{table}

The function of the gaze network is to generate a gaze map with a similar intensity distribution to the real gaze map, given drive-view images.
For quantitative evaluation, two standard metrics from saliency literature \cite{bylinskii2018different}, the Kullback-Leibler divergence (KL) and the Correlation Coefficient (CC), are adopted to evaluate the similarity. A larger similarity indicates better performance. A smaller KL divergence and a larger CC mean better similarity.
As the subject tends to look at the center area of the image, we also calculate the similarity between the image with a single Gaussian blob and the real gaze map as a comparison.

As shown in Table \ref{tab:gaze_pred}, for all the tracks, the estimated gaze map has better similarity with the real gaze map than the center Gaussian blob. For the two seen tracks (T1 and T2), the KL for the estimated gaze map is 75.6\% smaller than the center Gaussian blob, and the CC for the estimated gaze map is 23.6\% larger than that for the center Gaussian blob. For the unseen tracks (T3-T5), the KL for the estimated gaze map is 61.5\% smaller than that for the Gaussian blob, and the CC for the estimated gaze map is 9.4\% higher than that for the Gaussian blob on average.

\subsection{Gaze-Modulated Dropout Evaluation}
\subsubsection{Prediction error vs. drop probability}
To make a fair comparison, we scan over $dp$s from 0.1 to 0.8 with a step of 0.1 for both uniform dropout and gaze modulated dropout. The imitation networks are trained and tested in the same way (same $dp$ settings). Each image was forwarded into the imitation network 50 times with dropout.
The mean absolute estimation error between the steering angles generated by the network and the human driver are calculated each time. 

The average prediction error of the seen tracks stabilize at around three degrees for both dropout methods. The uniform dropout makes little difference while the gaze-modulated dropout has a significant effect in improving the model performance in unseen tracks. To be specify, prediction error of gaze modulated dropout shows a declining trend while for the uniform dropout, it rises slightly with the increase of $dp$.

We average the prediction error over all the tracks and choose the $dp$ corresponding to the smallest prediction errors. If not otherwise specified, $dp$ for gaze-modulated dropout is chosen to be 0.7 hereafter. For uniform dropout, it is 0.1. 

\subsubsection{Discussion on dropout implementation while testing}\label{sec:drop_imple}

As we performed the averaging in the aforementioned experiments, we noticed that the steering output varies while we stochastically drop out feature units. It is not reliable to directly apply gaze-modulated dropout to the network nor practicable to forward the input multiple times. Inspired by dropout, we directly multiply the features with the mask generated by the gaze map. In practice, we followed Algorithm\ref{alg:gaze_drop} but replaced the binary mask \textbf{M} with the keep probability mask \textbf{K} in line 7 while testing.

We compared the prediction error of the gaze-modulated dropout and uniform dropout with respect to different implementations.
As introduced in \cite{srivastava2014dropout}, forwarding the input without dropout at the testing stage approximates the averaging method. We compared the results of the averaging prediction error of multiple times forward passes and direct multiplying the keep probability pixel-wise as an approximation of the average shown in Fig. \ref{fig:error_total}. If we compare the first two and the last two groups of bars, both the uniform dropout and the gaze-modulated dropout show increases in prediction error for the ones approximating the averaging effect. The increment for gaze-modulated dropout is 0.08 degrees on average, which indicates the feasibility of multiplying the features with the gaze-generated mask in tests. 

\subsubsection{Performance on the dataset}\label{sec:pred_error}
\begin{figure}[!htb]
  \centering
    {\includegraphics[width=0.73\columnwidth]{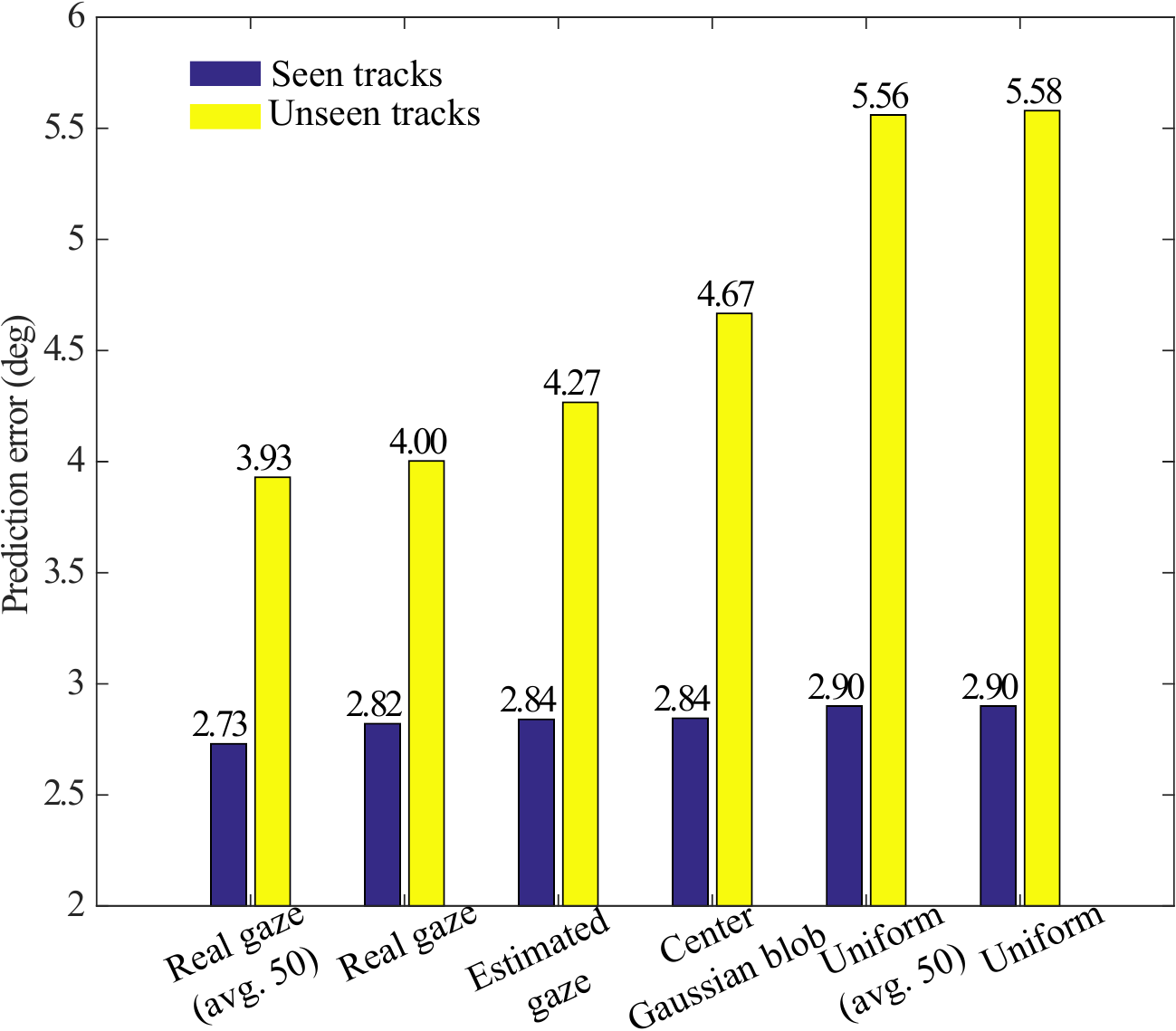}
     \caption{Quantitative performance on the dataset. Prediction errors of the model using real gaze, estimated gaze, center Gaussian blob modulated dropout and uniform dropout. \label{fig:error_total}}}
\vspace{-0.5em}     
\end{figure}

To evaluate the performance of the imitation network with modulated dropout, we first tested it on the testing dataset. Without loss of generality, we divided the tracks into seen tracks and unseen tracks and further averaged the prediction error of different tracks. $dp$ for estimated gaze and center Gaussian blob modulated dropout are set to 0.7.
As shown in Fig. \ref{fig:error_total}, for seen tracks, imitation networks with different dropout methods show close results. However, for unseen tracks, the network with estimated gaze-modulated dropout outperforms the uniform dropout network with 23.5\% lower prediction errors. The network with real gaze-modulated dropout achieves a 28.3\% decrease in the average estimation errors. Both networks also perform better than the network with center Gaussian blob dropout. On average, the imitation network with estimated gaze-modulated dropout achieves a 12.7\% decrease and the network with real gaze-modulated dropout achieves a 15.5\% decrease in the steering angle estimation error.

To summarize, either real or estimated gaze-modulated dropout improves the performance of the imitation networks in unseen environments and outperforms the center Gaussian blob modulated dropout and uniform dropout.

\subsubsection{Discussion on model uncertainty}\label{sec:model_uncertainty}
To identify the effectiveness of the imitation network from the perspective of uncertainty, we evaluate the model uncertainty of the imitation networks. We first scan over $dp$ from 0.1 to 0.9 with a step of 0.1. To match the two $dp$s for uniform dropout and gaze-modulated dropout, we recorded the drop rate of the gaze-modulated dropout for each sample and obtained an average drop probability across the testing data set.

\begin{figure}[!htb]
  \centering
    {\includegraphics[width=0.6\columnwidth]{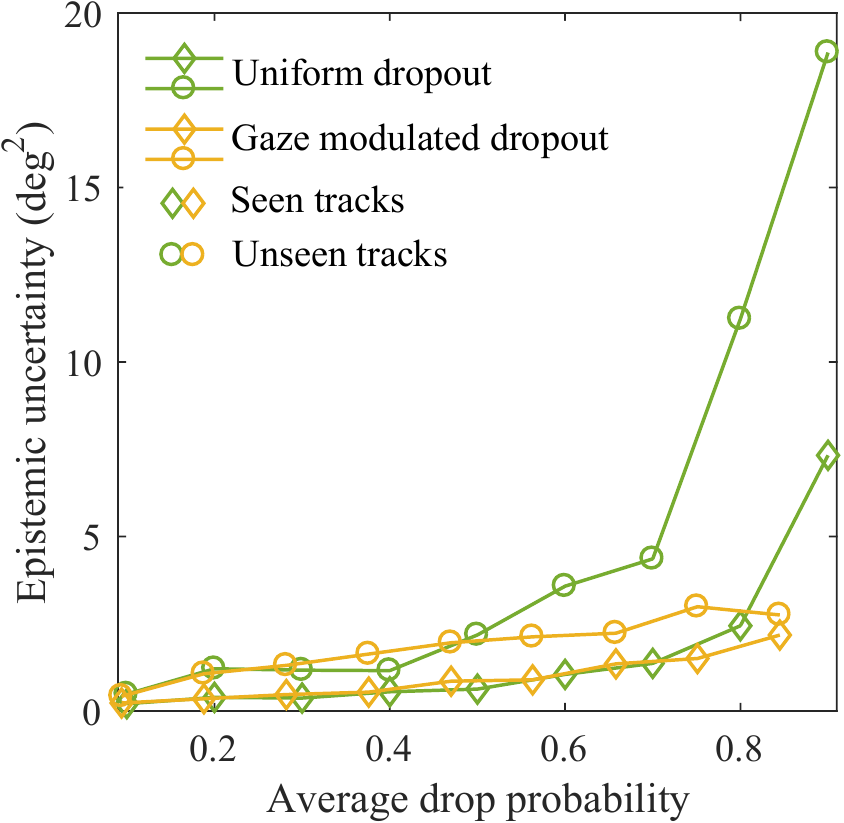}
     \caption{\textit{Epistemic} uncertainties of model using gaze-modulated dropout and uniform dropout with varied average drop probability. \label{fig:uncertainty_sweep}}}
     \vspace{-0.5em}
\end{figure}

As shown in Fig. \ref{fig:uncertainty_sweep}, it is clear to see that the model uncertainty for seen scenes is smaller than that of unseen scenes.
Also, we observe an increase in model uncertainty while increasing the average drop probability for both gaze-modulated dropout and uniform dropout, and seen and unseen scenes.
As the model uncertainties embody the familiarity of the model to the input images, we interpret results as that an unseen scene is less familiar and dropping more units will cause more loss of details so that the output features of dropout will vary more for multiple times forward passes. 

To compare the model uncertainties of gaze-modulated dropout with different dropout methods, the parameter $dp$ was chosen to ensure the same average drop probability. We trained models using uniform dropout with $dp$ equal to 0.66. Additionally, we also evaluate the model uncertainty for the center Gaussian blob modulated dropout and estimated gaze-modulated dropout with $dp$ equal to 0.7. We focused on the model uncertainties of unseen tracks. We found that gaze-modulated dropout and estimated gaze dropout had smaller model uncertainties, of 2.23 $deg^2$ and 2.60 $deg^2$ respectively. The center Gaussain blob and uniform dropout had larger model uncertainties, of 3.88 $deg^2$ and 4.36 $deg^2$ respectively.

Recall the prediction error evaluation in part \ref{sec:pred_error}, we found that the network with a lower model uncertainty achieves higher action estimation accuracy. This is consistent with the finding in \cite{kendall2016modelling}, where Kendall and Cipolla find that the model uncertainty is positively correlated with positional error for the camera relocalization task.
This indicates that the gaze-modulated dropout helps the network to capture the common features in unseen scenes and filter out the background area, which varies a lot over different scenes.

\subsection{Close loop performance}

We tested the performance of the networks in the simulator with unseen tracks. In each episode, the agent car started from a new location to drive along the path and overtake car if needed given the command from the human expert. The driving length of each episode was set to $2$ km. Human intervention will be given to drive the agent car until it returns to the road when infractions happen (e.g., driving outside the lane or collisions). We tested it in two cases: track with / without cars. In the case that no other cars are running on the road, the agent car simply follow the path. 

As shown in Table \ref{tab:simulator_test}, the proposed network with gaze-modulated dropout performs significantly better than the baseline network with uniform dropout. In terms of the success rate of cars overtaken, the proposed method is 31.5\% better than the baseline. In terms of average distance travelled before infractions and between two infractions, the proposed method is 55.2\% and 58.5\% better than baseline on average.

To see the performance comparison between the imitation networks with gaze-modulated dropout and uniform dropout running in the actual simulator, please refer to our supplementary video. The networks shown in the video were trained on the T1-T4 and test in T5 on the driving simulator.

\begin{table}[]
\begin{tabular}{llcc}
\hline
                                                                                                                   &                                                                                & With cars & No cars \\ \hline
\multirow{2}{*}{\begin{tabular}[c]{@{}l@{}}Success rate \\ of cars \\ overtaking (\%)\end{tabular}}                              & \begin{tabular}[c]{@{}l@{}}With uniform \\dropout\end{tabular}        & 67.6        & N/A           \\
                                                                                                                   & \begin{tabular}[c]{@{}l@{}}With estimated gaze\\modulated dropout\end{tabular} & \textbf{88.9}        & N/A           \\ \hline
\multirow{2}{*}{\begin{tabular}[c]{@{}l@{}}Ave. distance travelled \\from start  without \\ infractions (km)\end{tabular}}     & \begin{tabular}[c]{@{}l@{}}With uniform\\ dropout\end{tabular}        & 0.28        & 0.50           \\
                                                                                                                   & \begin{tabular}[c]{@{}l@{}}With estimated gaze \\modulated dropout\end{tabular} & \textbf{0.55}        & \textbf{0.57}           \\ \hline
\multirow{2}{*}{\begin{tabular}[c]{@{}l@{}}Ave. distance \\ travelled between \\ two infractions (km)\end{tabular}} & \begin{tabular}[c]{@{}l@{}}With uniform\\ dropout\end{tabular}        & 0.40        & 0.48           \\
                                                                                                                   & \begin{tabular}[c]{@{}l@{}}With estimated gaze \\modulated dropout\end{tabular} & \textbf{0.61}        & \textbf{0.79}          \\ \hline
\end{tabular}
\caption{Quantitative performance running in the testing track on the simulator. We compare the network with estimated gaze-modulated dropout and the network with uniform dropout. We measure the percentage of successful cars overtaking, average distance travelled without infractions (km) and average distance driven between infractions (km).}
\label{tab:simulator_test}
\vspace{-1em}
\end{table}


\section{Conclusions}\label{conclusion}
In this paper, we showed that a conditional GAN can generate an accurate estimation of human gaze maps. The learned gaze network generalizes well to unseen tracks. Furthermore, we proposed gaze-modulated dropout based on the generated gaze maps and incorporated it into the imitation network. We show that the use of gaze-modulated dropout significantly improves the human action estimation accuracy and decreases the model uncertainty. We demonstrated that a deep driving network could also benefit from expert's gaze as novice driver does. This work is an effort to take imitation learning to the next level by implicitly exploiting expert behavior.

Our work can be extended in several directions. First, the trained gaze network does not consider spatial-temporal characteristics of eye gaze. Including a spatio-temporal module such as recurrent net may improve the performance. Second, since the eye gaze data contains a wealth of information related to human intent, we may utilize the estimated gaze map to help choose the driving maneuvers, i.e., following and overtaking, which is currently done manually in our system. Finally, the proposed gaze-modulated dropout may also be applied to other tasks besides driving, such as vision-based robot navigation and robot manipulation.








\bibliographystyle{IEEEtran}
\bibliography{gazebib}
\end{document}